\pdfoutput=1
\PassOptionsToPackage{dvipsnames,table,HTML}{xcolor}
\documentclass{tjaa}

\usepackage[utf8]{inputenc}
\usepackage{lipsum}
\usepackage{amsmath}
\usepackage{multirow}
\usepackage{booktabs}
\usepackage{graphicx}
\usepackage{caption}
\usepackage{subcaption}
\usepackage{newtxtext}

\let\bibhang\relax

\usepackage[numbers,sort&compress]{natbib}
\newsavebox\verbbox

\title[Opportunistic Promptable Segmentation]{Opportunistic Promptable Segmentation: \newline Leveraging Routine Radiological Annotations to Guide 3D CT Lesion Segmentation}

\author[S. Church \others]{%
Samuel Church$^1$,
Joshua D. Warner$^2$,
Danyal Maqbool$^1$,
Xin Tie$^2$
\newauthor
Junjie Hu$^1$, Meghan G. Lubner$^2$, Tyler J. Bradshaw$^2$
\\
\adrid{1}University of Wisconsin--Madison, Department of Computer Sciences, Madison, WI USA\\
\adrid{2}University of Wisconsin--Madison, Department of Radiology, Madison, WI USA
}
\renewcommand{\pubyear}{2026}
\renewcommand{\volume}{0}
\renewcommand{\issue}{0}

\makeatletter
\def\@pubtype{Research Article}  
\def\@journal{}  
\def\@pubdoi{}
\makeatother
\begin{document}
\label{firstpage}
\pagerange{\pageref{firstpage}--\pageref{lastpage}}
\maketitle{M00-0000}

\begin{abstract}
The development of machine learning models for CT imaging depends on the availability of large, high-quality, and diverse annotated datasets. Although large volumes of CT images and reports are readily available in clinical picture archiving and communication systems (PACS), 3D segmentations of critical findings are costly to obtain, typically requiring extensive manual annotation by radiologists. On the other hand, it is common for radiologists to provide limited annotations of findings during routine reads, such as line measurements and arrows, that are often stored in PACS as grayscale softcopy presentation state (GSPS) DICOM objects. We posit that these sparse annotations can be extracted along with CT volumes and converted into 3D segmentations using promptable segmentation models, a paradigm we term Opportunistic Promptable Segmentation. To enable this paradigm, we propose SAM2CT, the first promptable segmentation model designed to convert radiologist annotations, including arrows and lines, into 3D segmentations in CT volumes. SAM2CT builds upon SAM2 by extending the prompt encoder to support arrow and line inputs and by introducing Memory-Conditioned Memories (MCM), a memory encoding strategy tailored to the characteristics of 3D medical volumes. On public lesion segmentation benchmarks, SAM2CT outperforms existing promptable segmentation models and similarly trained baselines, achieving Dice similarity coefficients of 0.649 for arrow prompts and 0.757 for line prompts. Applying the model to pre-existing GSPS annotations from a clinical PACS (N = 60), SAM2CT generates 3D segmentations that are clinically acceptable or require only minor adjustments in 87\% of cases, as scored by radiologists. Additionally, SAM2CT demonstrates strong zero-shot performance on select Emergency Department findings, including abscesses (DSC = 0.610) and gallstones (DSC = 0.725). These results suggest that large-scale mining of historical GSPS annotations represents a promising and scalable approach for generating 3D CT segmentation datasets.
\end{abstract}


\section{\Large Introduction}

\begin{figure*}
  \centering
  \begin{subfigure}{0.37\textwidth}
    \includegraphics[width=\linewidth]{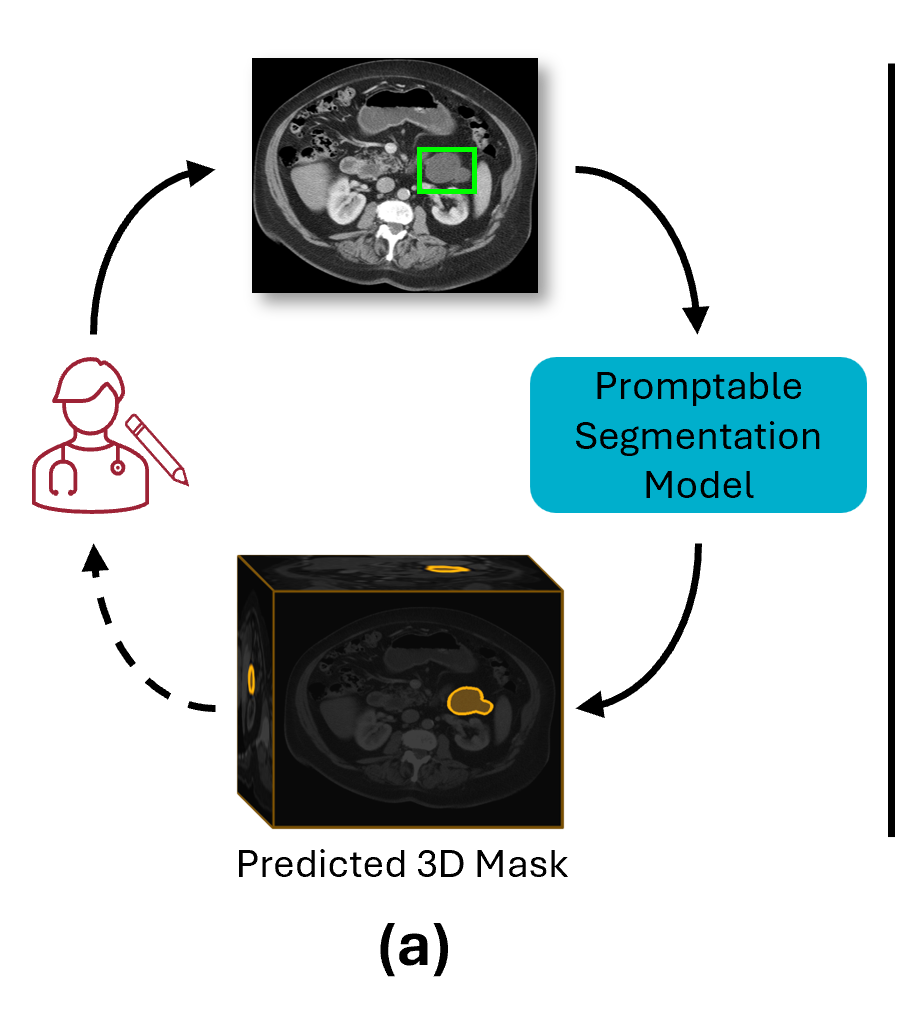}
    \phantomcaption
    \label{fig:ops_fig_a}
    
  \end{subfigure}%
  \begin{subfigure}{0.60\textwidth}
    \includegraphics[width=\linewidth]{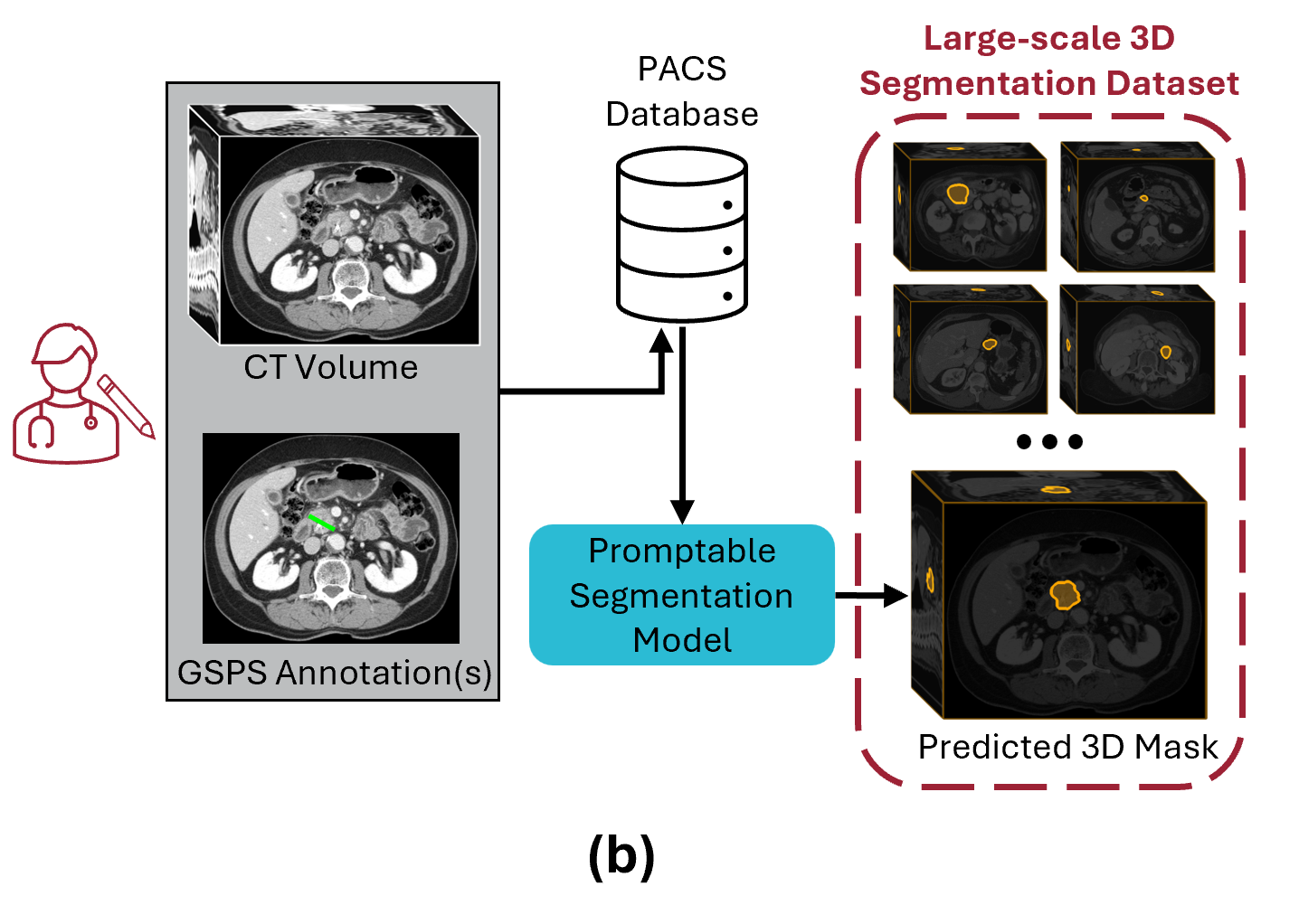}
    \phantomcaption
    \label{fig:ops_fig_b}
  \end{subfigure}
  \caption{ \normalsize
    \textbf{(a)} The standard interactive segmentation pipeline. A radiologist provides a prompt, which the promptable segmentation model converts to a 3D mask. Optionally, the radiologist can review this mask and provide corrections or additional prompts; \textbf{(b)} Our proposed opportunistic promptable segmentation  pipeline. Pre-existing GSPS annotations found in PACS are converted into 3D masks using a promptable segmentation model. This can be done automatically over large PACS databases to create large-scale 3D segmentation datasets.
  }
  \label{fig:ops_fig}
\end{figure*}
\begingroup
\renewcommand\thefootnote{}
\footnotetext{\normalsize Model/Code: \href{https://github.com/samdchurch/SAM2CT.git}{github.com/samdchurch/SAM2CT.git}}
\endgroup

\large
Artificial intelligence (AI) models are making an increasingly large impact in the field of radiology, as evidenced by regulatory approval for hundreds of new radiology AI devices in recent years \cite{fda_approval_jama, fda_approval_npj}. Many of these medical AI applications produce contours of findings within images, as these are needed for tasks such as visual grounding \cite{zach_visual_grounding}, grounded report generation \cite{maira_seg, organ_mask_guided_gen}, disease quantification \cite{tumor_volume_quantification}, and diagnosis \cite{brain_ct_classification, gen_ai_4_tumor_classification, panc_cancer_detection}. Development of these models fundamentally hinges upon the availability of large-scale contoured datasets, which can be difficult to obtain. This is particularly true for AI applications in CT, where 3D volumes are large, and relevant findings can be small, spatially dispersed, and highly variable in appearance. While some newer methods have utilized self-supervision to bypass the requirement of 3D segmentations \cite{ssl_dino_3d, 3d_ct_ssl}, supervised learning remains the gold-standard for reliable model development \cite{3D_transfer}.

\large
Acquiring CT volumes paired with 3D segmentations is a significant challenge. Most publicly available datasets primarily focus on the segmentation of organs \cite{amos_2022, ct-org, WORD-dataset} or specific tumor types \cite{lits_dataset, MSD-dataset, kits, uls23_dataset, lymph_node_dataset}. While healthcare centers often have tremendous numbers of CT scans available in their picture archiving and communication system (PACS), generating 3D segmentations of findings often requires manual contouring by radiologists – an expensive and time-consuming process. One promising alternative to dense manual annotations is interactive segmentation with promptable segmentation models. These models require simple prompts (e.g., bounding box) and then automatically convert those into 2D or 3D segmentations (Fig.~\ref{fig:ops_fig_a}). This greatly speeds up the data annotation pipeline as radiologists only need to provide general information about the object they want to segment, rather than the entire contour.

\large
Recent proliferation of promptable segmentation models largely stems from the Segment Anything Model (SAM) \cite{sam_paper}. 
SAM, which was trained on a massive dataset of 2D natural images and masks, interprets point prompts or bounding boxes and produces corresponding object masks. MedSAM \cite{medsam} expanded on SAM to allow segmentation of 2D medical image modalities, and other works extended SAM to allow for segmentation of 3D structures through 3D adaptors  \cite{sam_med_3d, sam3d}. The SAM2 model \cite{sam2} added a memory bank to the SAM architecture, allowing for segmentation of videos, which inherently extends to direct segmentation of 3D medical images without drastic architectural changes. Thus, several 3D medical image adaptations of SAM2 were subsequently developed which were fine-tuned on multiple public datasets across modalities \cite{medical_sam_2, MedSAM2}. Nonetheless, despite their potential to accelerate medical imaging annotation pipelines, these approaches remain fundamentally limited by their reliance on radiologists to generate the prompts.

\large
Alternatively, we posit that a vast number of radiologists-generated prompts for CT findings already exist, stored as DICOM grayscale softcopy presentation state (GSPS) objects inside of PACS. When a radiologist reviews a CT, they often make measurements of findings (e.g., diameter) using PACS-integrated tools, or place arrows to highlight key findings, allowing them to better track disease progression and communicate with referring providers. These measurements and arrows are typically stored as GSPS objects (or can be converted from proprietary formats into the standard GSPS format to allow for standardized exchange). These GSPS objects provide simple sparse information about the location of various CT findings without requiring additional physician involvement.

To take advantage of this wealth of prompts existing within PACS, we introduce a novel segmentation paradigm called Opportunistic Promptable Segmentation for CT imaging (Fig.~\ref{fig:ops_fig_b}). This paradigm, where GSPS annotations serve as prompts for automatic segmentation of CT findings, enables the creation of large-scale medical segmentation datasets without requiring any additional physician time beyond their standard clinical workflow. Since no current promptable CT segmentation models can interpret common GSPS annotations (i.e., line measurements and arrows), we developed our own promptable segmentation model, SAM2CT. SAM2CT extends SAM2 \cite{sam2} through fine-tuning on public lesion segmentation datasets with two key modifications: an updated prompt encoder capable of interpreting line and arrow prompts, and an enhanced memory pipeline. We evaluate SAM2CT on public lesion segmentation benchmarks as well as real GSPS annotations extracted from clinical PACS systems. Additionally, we show that our updated memory pipeline improves lesion segmentation  on other SAM2-based models during test-time with no further training required. To the best of our knowledge, this is the first model developed specifically for the conversion of GSPS annotations into 3D segmentations.

\section{\Large Materials and Methods}

\subsection{\large Dataset}

\large
We curated our training dataset from 5 public CT lesion segmentation datasets (Table~\ref{table:table_1_datasets}), covering a total of 8 tumor types \cite{lits_dataset, MSD-dataset, kits, uls23_dataset, lymph_node_dataset}. For external validation, we used the DeepLesion3D dataset \cite{deeplesion, uls23_dataset}, which contains 3D contours of abdominal and mediastinal lymph nodes; kidney, liver, and lung lesions; and a miscellaneous group of other lesion types. Due to a lack of bone lesions in other datasets, bone lesions found in the DeepLesion3D dataset were used in the training cohort rather than external validation to increase diversity of the training dataset.
To evaluate model performance on the opportunistic promptable segmentation task with real physician-provided data, we pulled 60 oncology CT exams from our clinical PACS that had GSPS annotations for lesions. With institutional review board approval, we selected 20 cases with arrow annotations, 20 with single line measurements, and 20 with major-minor axis measurements in a single plane. Additionally, we created an out-of-distribution test set of CT exams acquired in the Emergency Department. For case selection, we reviewed 30,000 Emergency Department reports and selected the 13 most common positive findings with corresponding GSPS annotations, using RadGraph to extract findings \cite{radgraph}. For each of these types of findings, 10 example findings from PACS, along with their GSPS object, were extracted and manually contoured, with review by two-board certified radiologists.

\begin{table}
    \normalsize
    \centering
    \caption{\normalsize Overview of public datasets used to train SAM2CT}
    \label{table:table_1_datasets}
    \begin{tabular}{|c|c|c|}
        \toprule
        Dataset Name & Lesion Category & train/val/test \\
        \midrule
        KiTS23 \cite{kits} & Kidney & 217/33/24 \\
        LiTS17 \cite{lits_dataset} & Liver & 454/44/46 \\
        \multirow{3}{*}{MSD \cite{MSD-dataset}} & Colon & 79/11/10 \\
        & Lung & 37/16/10 \\
        & Pancreas & 177/28/19 \\
        \multirow{2}{*}{NIH-LN \cite{lymph_node_dataset}} & Abd. Lymph Nodes & 376/139/42 \\
        & Med. Lymph Nodes & 251/100/24 \\
        DeepLesion3D \cite{deeplesion, uls23_dataset} & Bone & 71/13/13 \\
        \midrule
        All & - & 1662/384/188 \\
        \bottomrule
        
    \end{tabular}
\end{table}

\subsection{\large Prompt Generation}
\large 
Since public datasets lack corresponding GSPS-like annotations, we synthesized such annotations for both training and evaluation datasets. Given a 2D ground truth mask \textbf{G} selected from a lesion-containing axial slice, non-GSPS prompts (i.e., points, bounding boxes) were synthesized following the SAM2 protocol \cite{sam2}. We developed a similar approach for synthesizing arrow and line prompts.

To generate arrow annotations, we first computed the centroid of \textbf{G} and randomly selected an edge point on the mask boundary. Let $d$ denote the distance between the centroid and the selected edge point. The arrowhead was placed at a random location along the line connecting these two points, extended by up to $0.5d$ beyond the lesion boundary. The arrow tail was positioned at a fixed distance from the arrowhead and oriented toward the centroid, with a random angular perturbation of up to 15 degrees.

For line annotations, all possible lines connecting pairs of edge points on \textbf{G} were enumerated. During training, a line was randomly selected from the top 50th percentile of line lengths. During evaluation, selection was restricted to the top 90th percentile to better reflect typical clinical measurement behavior.

For training, \textbf{G} was sampled from all lesion-containing slices, with selection probability weighted by the lesion area in each slice. To more closely emulate radiologist-placed annotations during evaluation, \textbf{G} was restricted to the three axial slices with the largest lesion area.

\subsection{\large SAM2CT}
\begin{figure*}
  \normalsize
  \centering
  \includegraphics[width=1.0\textwidth]{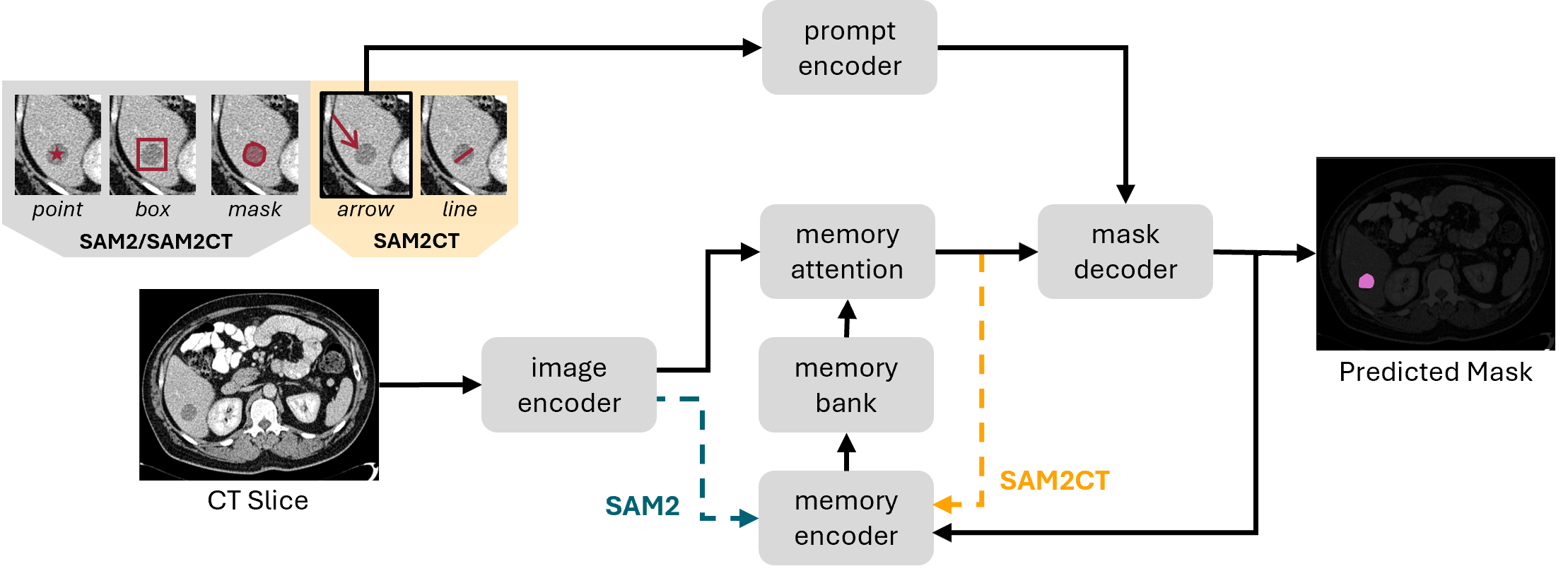}
  \caption {\normalsize Overview of SAM2CT Architecture. The blue arrow illustrates the standard SAM2 architecture where image features are fed directly from the image encoder to the memory encoder. The yellow arrow shows the memory-conditioned memories (MCM) of SAM2CT where the memory encoder receives image features after they have undergone memory attention.}
  \label{fig:SAM2CT_architecture}
\end{figure*}

\subsubsection{\large SAM2 Background}
\large
The Segment Anything Model (SAM) was originally developed for promptable segmentation of natural images using point and bounding-box prompts. SAM2 extends this framework to support segmentation of both images and videos. We selected SAM2 because it was trained on hundreds of millions of hours of video data across a diverse set of objects, and its architecture requires only minimal modification to support our opportunistic promptable segmentation task.

SAM2 consists of six main components: an image encoder, a prompt encoder, a memory attention module, a memory encoder, a memory bank, and a mask decoder. The model utilizes Hiera \cite{hiera}, a pretrained hierarchical vision transformer, as the image encoder. Given a frame at time $t$ from a video, $X_t$, the image encoder produces unconditioned image embeddings:

$$E_{unc}=\text{image\_encoder}(X_t)$$

Then, prompt inputs (i.e., points, bounding boxes, masks) are tokenized via the prompt encoder:

$$p_{tok}=\text{prompt\_encoder}(\text{prompt}_t)$$

For all frames prior to the current frame, $t$, SAM2's memory bank keeps track of n previous memory tokens, $mem_{t-n<i<t}$. Memory conditioned image embeddings are generated via memory attention between the unconditioned image embeddings and the memory bank:

$$E_{cond}=\text{memory\_attention}(E_{unc},mem_{t-n<i<t})$$

The mask decoder takes the memory conditioned image embeddings and the prompt tokens to generate a final predicted segmentation, $M_t$:

$$M_t=\text{mask\_decoder}(E_{cond},p_{tok})$$
Lastly, memory tokens for timestep t are stored in the memory bank using the predicted mask and the unconditioned image embeddings:

$$mem_t=\text{mask\_encoder}(E_{unc},M_t)$$
Thus, SAM2 can predict segmentations when given either a prompt (required for the initial frame) or memories from previous frames.

\subsubsection{\large Fine-tuning SAM2 on 3D Medical Data}
\large
SAM2 was designed for segmenting images and videos, but it can be used out-of-the-box for 3D medical images, such as CTs and MRIs, by segmenting slice-by-slice. Given a prompt on slice $k$, SAM2 can create a 3D prediction via an inferior processing step and a superior processing step. For both steps, SAM2 starts at slice $k$ and segmentation will halt early if SAM2 predicts an empty mask for a given axial slice.
First pass, starting on slice $k$ and increasing,
$$M[k]=\text{SAM2}(X_t ),k\leq t \leq N$$
Second pass, starting on slice $k$ and decreasing,
$$M[k]=\text{SAM2}(X_t ),k\geq t \geq0$$
SAM2 was not originally trained to produce high-quality segmentations of medical structures. To enable accurate lesion segmentation, SAM2 must therefore be fine-tuned on CT data. To reduce computational cost, fine-tuning is performed on 8-slice sub-volumes sampled from each CT scan, rather than processing the full volume at every training epoch.

\subsubsection{\large SAM2CT Architecture}
\large
Fine-tuning SAM2 on 3D medical data is an effective method, but architectural changes to the memory encoding process can further increase segmentation performance. Specifically, memories in SAM2 are formed using the output mask and the unconditioned image embeddings. The unconditioned image embeddings do not contain information about where the specific object of interest is. While this approach may be useful in the video segmentation setting where tracking information about the entire scene is important for handling challenges such as occlusion, in the volumetric segmentation setting, it is important to encode memories that are as specific to the target lesion as possible. Because of this, we created SAM2CT (Fig.~\ref{fig:SAM2CT_architecture}), a fine-tuned SAM2 model that creates memories by fusing the predicted mask with the conditioned image embeddings,
$$mem_{t}=\text{mask\_encoder}(E_{cond},M_t)$$

These memory-conditioned memories (MCM) express more lesion-specific information, which then goes on to provide better segmentations on future slices.

Additionally, SAM2CT expands SAM2’s prompt library to accept arrow and line prompts. To incorporate this capability, we added three additional learnable parameters within the prompt encoder: line endpoint, arrow start, and arrow end. When passing a line measurement prompt, the prompt encoder receives each line endpoint added with the line endpoint learnable parameter. Similarly, arrows consist of the arrow start point and arrow endpoint summed with their respective learnable parameter.

\subsubsection{\large SAM2CT Training}
\large
SAM2CT was fine-tuned end-to-end initialized with SAM2's tiny model (38.9M parameters). As suggested in previous work \cite{MedSAM2}, we also found that model size did not have a significant impact on segmentation accuracy. We trained SAM2 on 8-slice batches for 80 epochs. The best model checkpoint was selected using the validation set. The base learning rate and image encoder learning rate were $5e^{-6}$ and $3e^{-6}$, respectively. Image values were clipped at -500 and 500 HU, regardless of lesion type. Subsequently, images were converted to 8-bit RGB to conform to SAM2’s input specifications.

\subsection{\large Comparator Methods}
\large
To ensure a fair evaluation of the proposed SAM2CT architecture, we implemented a variant of SAM2 that includes support for arrow- and line-based prompts but excludes the MCM architecture present in SAM2CT. This model, referred to as SAM2 (FT), was trained and evaluated using the same data, optimization settings, and evaluation protocol as SAM2CT.

Because no off-the-shelf segmentation models natively accept arrow- or line-style prompts, we trained two additional comparison models from scratch. We selected two representative UNet-derived architectures with established strong performance on medical image segmentation tasks. The first is DynUNet, MONAI's implementation of nnU-Net \cite{nnunet, monai}, which serves as a fully-convolutional baseline. The second is Swin UNETR \cite{swin_unetr}, which integrates a Swin-transformer backbone with the UNet decoder and represents a hybrid convolution–transformer approach. To provide these models with prompt information, prompts were encoded as binary masks and appended to the image input as a second channel with the same spatial dimensions as the image. All models were trained for 150 epochs, utilizing a 20-epoch linear warmup phase followed by a cosine annealing learning rate schedule. The peak learning rate was set to $5e^{-3}$. Optimization was performed using the AdamW optimizer to minimize a combination loss function consisting of Dice loss and Focal loss.

\begin{figure*}
  \normalsize
  \centering
  \includegraphics[width=1.0\textwidth]{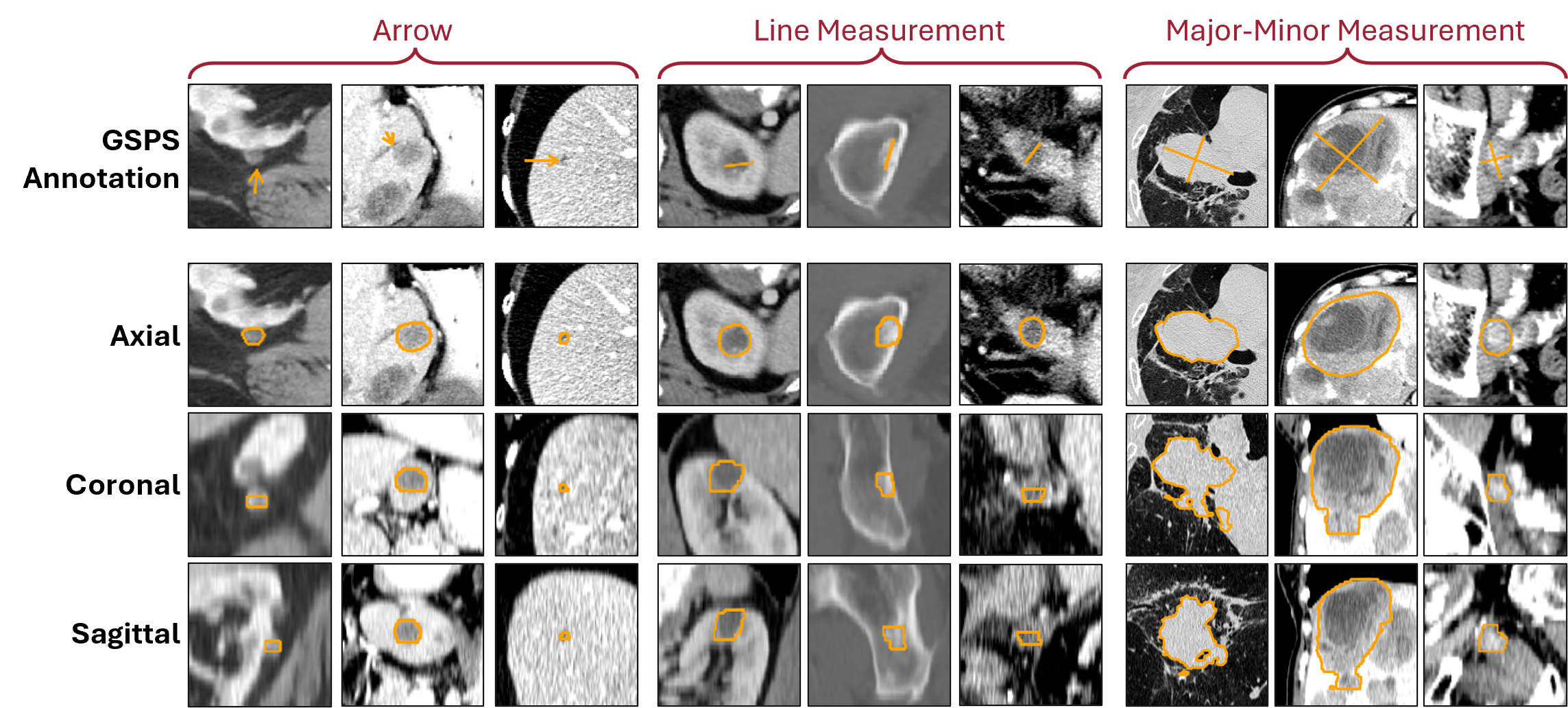}
  \caption {\normalsize Visualization of SAM2CT produced contours for oncology cases. Original GSPS prompts (arrow, line measurement, and major-minor measurements) are shown alongside SAM2CT segmentations in axial, coronal, and sagittal views. }
  \label{fig:examples_fig}
\end{figure*}

\subsection{\large Evaluation Metrics}
\large
Model performance was evaluated using Dice similarity coefficient (DSC) and the percent difference of response evaluation criteria in solid tumors (RECIST) measurements \cite{recist}. For all results, 95\% confidence intervals were calculated using bootstrap resampling. 

The DSC, defined as 
$$DSC=\frac{2|A \cap B|}{|A|+|B|}$$
quantifies volumetric overlap between predicted (A) and reference (B) segmentations.

To evaluate a metric with more direct clinical significance, we also report the average percent difference RECIST 1.1 score between the predicted and ground-truth segmentations. RECIST 1.1 measures the longest oblique axial measurement contained within the lesion. Given ground-truth and predicted RECIST measurement ($m_{gt}$ and $m_{pred}$, respectively) for an individual lesion, the percent difference is:

$$100*\frac{|m_{pred}-m_{gt}|}{m_{gt}}$$

For evaluation on real-world oncology cases with PACS annotations, two board-certified radiologists independently reviewed the predicted segmentation masks. Each prediction was assigned a score on a three-point scale: scores of 3 indicated masks acceptable for clinical measurements without modification, scores of 2 indicated masks requiring minor adjustments, and scores of 1 indicated masks requiring major adjustments to be suitable for clinical measurements.  

\begin{table*}
    \centering
    \normalsize
    \caption{\normalsize Dice scores are reported with 95\% confidence intervals for all models. Results are stratified by dataset and prompt type.}
    \label{table:main_results}
    \makebox[0pt][c]{

        \begin{tabular}{|c|l|c|c|c|c|c|}
            \toprule
            Dataset & Model & Arrow & Line & Point & BBox & Mask \\
            \midrule
            \multirow{6}{*}{MSD Colon} & SAM2 & - & - & 0.131 [0.080,0.186] & 0.307 [0.259,0.359] & 0.419 [0.348,0.493] \\
             & MedSAM2 & - & - & - & 0.635 [0.580,0.686] & - \\
             & DynUNet & 0.414 [0.328,0.495] & 0.581 [0.522,0.636] & 0.457 [0.388,0.528] & 0.556 [0.503,0.610] & 0.605 [0.558,0.647] \\
             & Swin UNETR & 0.345 [0.285,0.407] & 0.559 [0.506,0.604] & 0.421 [0.368,0.472] & 0.550 [0.510,0.589] & 0.595 [0.553,0.633] \\
             & SAM2 (FT) & 0.627 [0.552,0.697] & 0.670 [0.599,0.735] & 0.626 [0.549,0.699] & 0.711 [0.649,0.764] & 0.754 [0.708,0.796] \\
             & SAM2CT & \textbf{0.675 [0.612,0.733]} & \textbf{0.708 [0.652,0.759]} & \textbf{0.653 [0.587,0.717]} & \textbf{0.729 [0.677,0.773]} & \textbf{0.760 [0.717,0.799]} \\
            \midrule
    
            \multirow{6}{*}{MSD Lung} & SAM2 & - & - & 0.272 [0.166,0.387] & 0.501 [0.414,0.587] & 0.419 [0.348,0.493] \\
             & MedSAM2 & - & - & - & 0.717 [0.651,0.776] & - \\
             & DynUNet & 0.354 [0.268,0.439]	& 0.588 [0.537,0.636] & 0.496 [0.412,0.570] & 0.662 [0.622,0.700]	& 0.709 [0.664,0.747] \\
             & Swin UNETR & 0.286 [0.195,0.382] & 0.472 [0.388,0.555] & 0.372 [0.283,0.462] & 0.480 [0.403,0.555]	& 0.627 [0.576,0.678] \\
             & SAM2 (FT) & 0.645 [0.546,0.731] & \textbf{0.758 [0.713,0.797]}	& 0.677 [0.588, 0.751] & \textbf{0.759 [0.713,0.798]} & 0.792 [0.741,0.833] \\
             & SAM2CT & \textbf{0.674 [0.583,0.752]} & \textbf{0.756 [0.696,0.804]} & \textbf{0.698 [0.620,0.763]} & 0.751 [0.690,0.798] & \textbf{0.797 [0.742,0.840]} \\
            \midrule
    
            \multirow{6}{*}{MSD Pancreas} & SAM2 & - & - & 0.135 [0.080,0.197] & 0.408 [0.343,0.476] & 0.511 [0.447,0.576] \\
             & MedSAM2 & - & - & - & 0.746 [0.703,0.784] & - \\
             & DynUNet & 0.442 [0.382,0.501]	& 0.668 [0.629,0.705]	& 0.585 [0.534,0.633] & 0.742 [0.717,0.767]	& 0.763 [0.735,0.790] \\
             & Swin UNETR & 0.407 [0.350,0.465] & 0.600 [0.555,0.644] & 0.550 [0.503,0.595] & 0.661 [0.623,0.698]	& 0.693 [0.660,0.724]\\
             & SAM2 (FT) & \textbf{0.703 [0.646,0.757]} & 0.752 [0.709,0.790]	& 0.735 [0.688,0.777] & 0.796 [0.767,0.822]	& 0.826 [0.801,0.850] \\
             & SAM2CT & 0.697 [0.634,0.754] & \textbf{0.773 [0.744,0.801]} & \textbf{0.744 [0.704,0.782]} & \textbf{0.816 [0.795,0.836]}	& \textbf{0.835 [0.813,0.856]} \\
            \midrule
    
            \multirow{6}{*}{KiTS23} & SAM2 & - & - & 0.305 [0.231,0.381] & 0.565 [0.506,0.625] & 0.634 [0.579,0.688] \\
             & MedSAM2 & - & - & - & \textbf{0.866 [0.846,0.885]} & - \\
             & DynUNet & 0.420 [0.355,0.486]	& 0.689 [0.650,0.727] & 0.553 [0.487,0.617] & 0.779 [0.753,0.804]	& 0.825 [0.800,0.848] \\
             & Swin UNETR & 0.249 [0.200,0.301] & 0.544 [0.491,0.597] & 0.430 [0.375,0.486] & 0.606 [0.553,0.657]	& 0.682 [0.636,0.724]\\
             & SAM2 (FT) & \textbf{0.821 [0.780,0.856]} & \textbf{0.850 [0.817,0.877]}	& 0.811 [0.764,0.850] & \textbf{0.868 [0.843,0.889]}	& \textbf{0.880 [0.852,0.904]} \\
             & SAM2CT & \textbf{0.822 [0.780,0.857]} & \textbf{0.850 [0.817,0.878]} & \textbf{0.821 [0.778,0.857]} & \textbf{0.868 [0.846,0.889]} & \textbf{0.878 [0.851,0.901]} \\
            \midrule
    
            \multirow{6}{*}{LiTS17} & SAM2 & - & - & 0.068 [0.034,0.108] & 0.460 [0.414,0.507] & 0.525 [0.475,0.574] \\
             & MedSAM2 & - & - & - & 0.717 [0.680, 0.751] & - \\
             & DynUNet & 0.220 [0.184,0.257]	& 0.559 [0.527,0.591] & 0.381 [0.340,0.421] & 0.636 [0.608,0.664]	& 0.684 [0.654,0.712] \\
             & Swin UNETR & 0.194 [0.164,0.224] & 0.334 [0.298,0.368] & 0.239 [0.209,0.270] & 0.321 [0.286,0.356]	& 0.499 [0.457,0.540] \\
             & SAM2 (FT) & 0.671 [0.622,0.717] & \textbf{0.778 [0.752,0.801]}	& 0.742 [0.705,0.776] & 0.791 [0.771,0.810]	& 0.844 [0.828,0.860] \\
             & SAM2CT & \textbf{0.676 [0.627,0.721]} & \textbf{0.775 [0.747,0.801]} & \textbf{0.757 [0.727,0.785]} & \textbf{0.799 [0.781,0.816]} & \textbf{0.841 [0.824,0.857]} \\
            \midrule
    
            \multirow{6}{*}{NIH-ABD} & SAM2 & - & - & 0.054 [0.028,0.083] & 0.277 [0.233,0.322] & 0.341 [0.296,0.387] \\
             & MedSAM2 & - & - & - & 0.658 [0.620,0.694] & - \\
             & DynUNet & 0.226 [0.195,0.258] & 0.553 [0.529,0.578] & 0.448 [0.417,0.478] & 0.615 [0.595,0.635] & 0.661 [0.641,0.680] \\
             & Swin UNETR & 0.399 [0.367,0.433] & 0.470 [0.440,0.499] & 0.422 [0.389,0.455] & 0.498 [0.468,0.529]	& 0.579 [0.549,0.608] \\
             & SAM2 (FT) & 0.547 [0.494,0.600] & 0.679 [0.644,0.711]	& 0.653 [0.613,0.691] & 0.700 [0.668,0.731]	& 0.744 [0.715,0.770] \\
             & SAM2CT & \textbf{0.565 [0.509,0.618]} & \textbf{0.706 [0.674,0.735]} & \textbf{0.663 [0.623,0.702]} & \textbf{0.732 [0.704,0.759]} & \textbf{0.753 [0.727,0.778]} \\
            \midrule
    
            \multirow{6}{*}{NIH-MED} & SAM2 & - & - & 0.028 [0.011,0.051] & 0.195 [0.160,0.230] & 0.333[0.293,0.373] \\
             & MedSAM2 & - & - & - & 0.647 [0.597,0.693] & - \\
             & DynUNet & 0.222 [0.179,0.265]	& 0.591 [0.564,0.616] & 0.471 [0.436,0.506] & 0.652 [0.632,0.671]	& 0.677 [0.657,0.696] \\
             & Swin UNETR & 0.402 [0.351,0.452] & 0.533 [0.497,0.568] & 0.448 [0.403,0.492] & 0.534 [0.490,0.574]	& 0.611 [0.583,0.637] \\
             & SAM2 (FT) & 0.621 [0.573,0.667] & 0.683 [0.650,0.716]	& 0.656 [0.615,0.694] & 0.715 [0.687,0.742]	& \textbf{0.766 [0.743,0.787]} \\
             & SAM2CT & \textbf{0.666 [0.626,0.703]} & \textbf{0.718 [0.687,0.747] } & \textbf{0.668 [0.627,0.705]} & \textbf{0.739 [0.714,0.763]}	& \textbf{0.766 [0.743,0.787]} \\
            \midrule
    
            \multirow{6}{*}{DL3D-Bone} & SAM2 & - & - & 0.103 [0.034,0.184] &	0.516 [0.444,0.589]	& 0.546 [0.471,0.622] \\
             & MedSAM2 & - & - & - & 0.517 [0.427,0.601] & - \\
             & DynUNet & 0.362 [0.293,0.433]	& 0.612 [0.559,0.662] & 0.420 [0.351,0.494] & 0.655 [0.609,0.698]	& 0.723 [0.677,0.762] \\
             & Swin UNETR & 0.411 [0.328,0.492] & 0.460 [0.392,0.527 & 0.407 [0.318,0.492] & 0.424 [0.341,0.508]	& 0.570 [0.507,0.633] \\
             & SAM2 (FT) & 0.405 [0.298,0.516] & \textbf{0.765 [0.721,0.805]} & 0.748 [0.689,0.799] & 0.767 [0.722,0.807]	& \textbf{0.802 [0.768,0.833]} \\
             & SAM2CT & \textbf{0.422 [0.313,0.530]} & \textbf{0.768 [0.724,0.808]} & \textbf{0.764 [0.719,0.805]} & \textbf{0.779 [0.737,0.817]} & 0.792 [0.748,0.829] \\
            \midrule
            \midrule
    
            \multirow{6}{*}{All Datasets} & SAM2 & - & - & 0.137 [0.115,0.160] & 0.404 [0.383,0.424]	& 0.492 [0.470,0.514] \\
             & MedSAM2 & - & - & - & 0.688 [0.669,0.706] & - \\
             & DynUNet & 0.370 [0.339,0.400]	& 0.617 [0.597,0.637] & 0.494 [0.467,0.523] & 0.675 [0.659,0.691]	& 0.717 [0.702,0.732] \\
             & Swin UNETR & 0.296 [0.269,0.325] & 0.502 [0.477,0.527] & 0.402 [0.376,0.428] & 0.524 [0.501,0.545]	& 0.619 [0.600,0.628] \\
             & SAM2 (FT) & 0.630 [0.605,0.653] & 0.742 [0.727,0.756]	& 0.706 [0.687,0.724] & 0.763 [0.750,0.776]	& \textbf{0.801 [0.790,0.812]} \\
             & SAM2CT & \textbf{0.649 [0.626,0.672]} & \textbf{0.757 [0.743,0.770]} & \textbf{0.721 [0.704,0.738]} & \textbf{0.777 [0.764,0.788]} & \textbf{0.803 [0.791,0.813]} \\
            \bottomrule
        \end{tabular}
    }
\end{table*}

\begin{table*}
    \centering
    \normalsize
    \caption{\normalsize Dice score results for bounding box prompt on DeepLesion3D external data.}
    \label{table:external_test_results}
    \makebox[0pt][c]{
    \begin{tabular}{|l|c|c|c|c|c|c|}
    \toprule
    Model & Kidney & Liver & Lung & Abd-LN & Med-LN & Misc. \\
    \midrule
    SAM2 & 0.374 [0.291,0.459] & 0.503 [0.411,0.593] & 0.409 [0.364,0.455] &	0.400 [0.358,0.443]	& 0.328 [0.283,0.376] & 0.541 [0.484,0.599] \\
    MedSAM2 & 0.566 [0.484,0.643] & 0.552 [0.436,0.664]	& \textbf{0.586 [0.528,0.640]} & 0.442 [0.392,0.491]	& 0.565 [0.518,0.611] & 0.620 [0.556,0.682] \\
    SAM2 (FT) & 0.708 [0.644,0.764]	& 0.750 [0.705,0.791] & 0.532 [0.476,0.588]	& 0.690 [0.654,0.725] & 0.728 [0.699,0.756]	& 0.768 [0.723,0.807] \\
    SAM2CT & \textbf{0.720 [0.664,0.771]} & \textbf{0.773 [0.735,0.811]} & 0.547 [0.494,0.600] & \textbf{0.704 [0.672,0.734]} & \textbf{0.734 [0.702,0.764]} & \textbf{0.776 [0.735,0.813]} \\
    \bottomrule
    
    \end{tabular}
    }
\end{table*}

\begin{table}
    \centering
    \normalsize
    \caption{\normalsize Percent error of RECIST measurements on hold-out test set and external test set (DeepLesion3D)}
    \label{table:recist_results}
    \begin{tabular}{|l|c|c|}
        \toprule
        & \multicolumn{2}{c|}{RECIST Error (\%)} \\
        \cmidrule{2-3}
        Model & Hold-out Test Set & External Test Set \\
        \midrule
        SAM2        & 85.6 [54.7,126.]   & 94.0 [64.3,135.] \\
        MedSAM2     & 25.6 [16.4,36.4]   & 50.8 [37.4,67.9] \\
        SAM2 (FT)   & 12.0 [9.71,13.9]   & 16.9 [9.13,28.1] \\
        SAM2CT      & \textbf{11.1 [9.13,12.9]}   & \textbf{15.9 [8.49,26.5]} \\
        \bottomrule
    \end{tabular}
\end{table}

\section{\Large Results}
\subsection{\large Hold-out Test Set}
\large
As shown in Table~\ref{table:main_results}, on the hold-out test set, SAM2CT achieved the highest Dice scores across all five prompt types. The Dice scores for SAM2CT were 0.649, 0.757, 0.721, 0.777, and 0.803 for arrow, line, point, box, and mask prompts, respectively. Across prompt types, SAM2CT consistently outperformed the other models, followed by SAM2 (FT), MedSAM2 (box), DynUNet, Swin UNETR, and then SAM2 (point, box, mask).

In addition to segmentation accuracy, we evaluated models based on their performance in producing RECIST measurements for individual lesions (Table~\ref{table:recist_results}). SAM2CT demonstrated the lowest average RECIST error at 11.1\%, compared with 12.0\% for SAM2 (FT), 25.6\% for MedSAM2, and 85.6\% for SAM2.

\subsection{\large External Test Set}
\large
We further assessed model generalization on the DeepLesion3D dataset (Table~\ref{table:external_test_results}). SAM2CT achieved the highest Dice scores for five of the six lesion types: Kidney (0.720), Liver (0.773), Abdominal Lymph Node (0.704), Mediastinal Lymph Node (0.734), and Miscellaneous Lesions (0.776). MedSAM2 outperformed SAM2CT on lung tumors, with Dice scores of 0.586 and 0.547, respectively.

For RECIST error on individual DeepLesion3D lesions (Table~\ref{table:recist_results}), SAM2CT again achieved the lowest average error at 15.9\%, followed by SAM2 (FT) at 16.9\%, MedSAM2 at 50.8\%, and SAM2 at 94

\subsection{\large Real-World Oncology and Emergency Department Cases}
\large
Among the 60 oncology cases containing GSPS annotations (20 arrow prompts, 20 line prompts, and 20 major–minor axis measurements), radiologists found that 62\% of SAM2CT-generated masks required no adjustments, 25\% of cases required minor adjustments, and 13\% required major adjustments to achieve clinical acceptability. Arrow-based annotations were the most likely to require either minor or major refinement. Examples of 3D segmentations made by SAM2CT are shown in Fig.~\ref{fig:examples_fig}.

For Emergency Department (ED) cases (Table~\ref{table:ed_results}), SAM2CT achieved an average Dice similarity coefficient (DSC) of 0.581 across all findings when using the original GSPS prompt, compared to 0.567 for SAM2 (fine-tuned). When provided with bounding box prompts, SAM2CT and SAM2 (fine-tuned) performed similarly (DSC = 0.667 and 0.665, respectively), while MedSAM2 achieved a slightly lower DSC of 0.659. When using the original clinical annotations, SAM2CT exhibited substantial variability across findings, achieving higher performance for abscesses (DSC = 0.610), cysts (DSC = 0.727), and gallstones (DSC = 0.725), while performance was notably lower for diverticulitis  (DSC = 0.302), hernias (DSC = 0.371), and kidney stones (DSC = 0.419).

\begin{figure}
    \centering
    \includegraphics[width=\columnwidth]{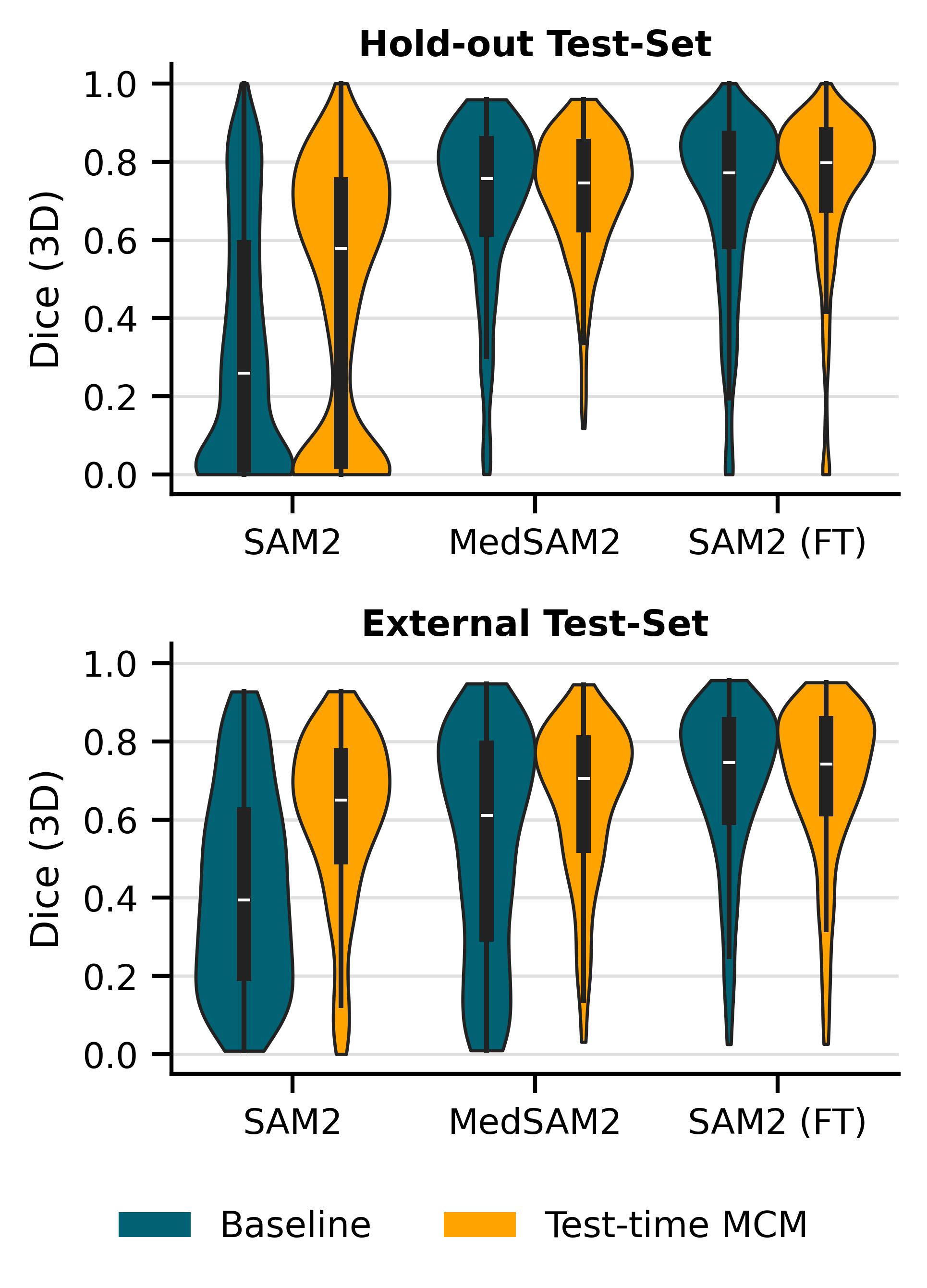}
    \caption{\normalsize Performance of models with test-time memory-conditioned-memory (MCM) model architecture adaptation. Results are shown for the hold-out test set (top) and the external test set (bottom).}
    \label{fig:test_time_mcm}
\end{figure}

\subsection{\large Test-time MCM}
\large
We evaluated the impact of adapting SAM-based architectures to use MCM at test time, without additional fine-tuning. We evaluated three models: SAM2, MedSAM2, and SAM2 (FT). Results for these three models on both internal and external data can be seen in Fig.~\ref{fig:test_time_mcm}. For SAM2, applying MCM led to a substantial improvement in Dice scores, increasing performance by 0.121 on the hold-out test set (p < 0.001) and by 0.172 on the DeepLesion3D external dataset (p = 0.031). For MedSAM2, MCM produced smaller gains: a 0.021 increase on the hold-out test set (p = 0.547) and a 0.097 increase on the external dataset (p = 0.031), indicating limited but statistically significant improvements on the external data. For SAM2 (FT), MCM increased Dice scores by 0.040 on the hold-out test set (p < 0.001) and 0.007 on the external dataset (p = 0.1562). Notably, this brought SAM2 (FT) + MCM performance very close to SAM2CT, with a hold-out test Dice score of 0.738 versus 0.741 for SAM2CT, and an external dataset Dice score of 0.703 versus 0.709 for SAM2CT.

\begin{table*}
    \normalsize
    \centering
    \caption{\normalsize Dice score results on Emergency Department data. Results are reported using the original annotations (i.e., combination of arrows, single line measurements, major-minor axes measurements) and generated bounding box annotations}
    \label{table:ed_results}
    \begin{tabular}{lcccccc}

    \toprule
     & \multicolumn{2}{c}{Original Annotations} & & \multicolumn{3}{c}{Box Annotations} \\
    \cmidrule(lr){2-3} \cmidrule(lr){5-7}

     & SAM2CT & SAM2 (FT) & & SAM2CT & SAM2 (FT) & MedSAM2 \\
     \midrule
     Abscess	& 0.610 [0.520,0.703] & 0.554 [0.416,0.680] & & 0.624 [0.499,0.730] & 0.620 [0.498,0.730] & 0.685 [0.545,0.785] \\
     Aneurysm & 0.583 [0.430,0.724] & 0.524 [0.369,0.673] & & 0.579 [0.426,0.719] & 0.528 [0.367,0.682] & 0.508 [0.372,0.646] \\
     Appendicitis & 0.540 [0.408,0.667] & 0.549 [0.415,0.678] & & 0.600 [0.492,0.700] & 0.624 [0.513,0.718]	& 0.580 [0.463,0.688] \\
     Cyst & 0.727 [0.536,0.875] & 0.725 [0.534,0.871] & & 0.793 [0.680,0.884] & 0.790 [0.677,0.883] & 0.774 [0.609,0.885] \\
     Diverticulitis & 0.302 [0.174,0.421] & 0.295 [0.122,0.487] & & 0.480 [0.366,0.599] & 0.479 [0.358,0.603]	& 0.629 [0.525,0.735] \\
     Gallstone & 0.725 [0.543,0.852]	& 0.741 [0.563,0.856] & & 0.830 [0.780,0.879] & 0.829 [0.784,0.873]	& 0.653 [0.466,0.824] \\
     Hematoma & 0.528 [0.269,0.767] & 0.514 [0.281,0.732] & & 0.713 [0.536,0.839] & 0.706 [0.576,0.799] & 0.790 [0.717,0.856] \\
     Hernia & 0.371 [0.186,0.558] & 0.345 [0.178,0.518] & & 0.550 [0.387,0.716] & 0.572 [0.393,0.739] & 0.724 [0.586,0.840] \\
     Kidney Stone & 0.419 [0.239,0.601] & 0.387 [0.201,0.587] & & 0.570 [0.407,0.709] & 0.584 [0.427,0.713]	& 0.420 [0.261,0.567] \\
     Lesion & 0.683 [0.602,0.761] & 0.681 [0.600,0.762] & & 0.730 [0.684,0.777] & 0.729 [0.683,0.784] & 0.643 [0.484,0.758] \\
     Lymph Node & 0.736 [0.692,0.783] & 0.742 [0.693,0.791] & & 0.783 [0.748,0.825] & 0.779 [0.728,0.827]	& 0.746 [0.680,0.799] \\
     Mass & 0.683 [0.512,0.816] & 0.689 [0.543,0.817] & & 0.786 [0.717,0.843] & 0.788 [0.719,0.845] & 0.785 [0.703,0.855] \\
     Nodule & 0.588 [0.431,0.711] & 0.574 [0.425,0.692] & & 0.639 [0.474,0.753] & 0.622 [0.462,0.738]	& 0.666 [0.599,0.734] \\
     \midrule
     Average	& 0.581 [0.532,0.626] & 0.567 [0.518,0.615] &	& 0.667 [0.629,0.705] & 0.665 [0.627,0.701] & 0.659 [0.619,0.696] \\
    \bottomrule

    \end{tabular}
\end{table*}

\begin{figure*}
  \centering
  \includegraphics[width=1.0\textwidth]{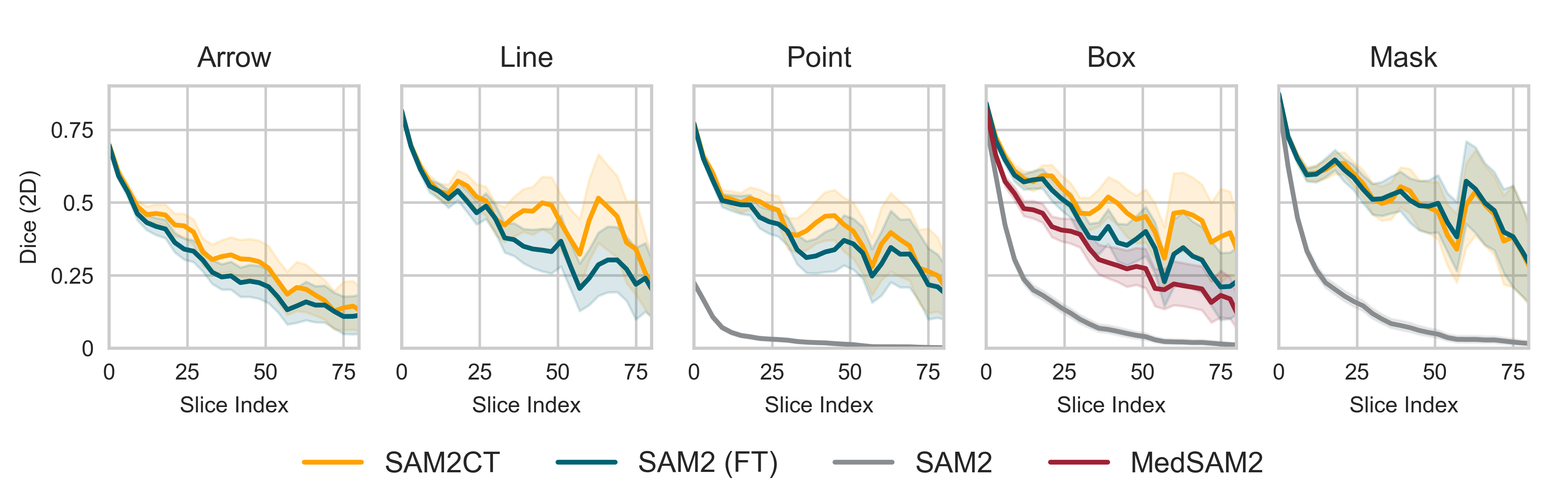}
  \caption {\normalsize 2D Dice scores on hold-out test set as a function of out-of-plane distance from the original prompt. Slice index 0 corresponds to the prompted slice.}
  \label{fig:slice_dices}
\end{figure*}

\section{\Large Discussion}
\large
Previous work has illustrated the potential of physician-in-the-loop promptable segmentation models for 3D medical images \cite{MedSAM2}; however, SAM2CT offers the first model to address opportunistic promptable segmentation. We successfully adapted the SAM2 architecture so it can interpret the annotations created by physicians as part of their routine workflow, namely line measurements and arrows, and convert them into 3D segmentations. This creates an avenue for large-scale segmentation dataset generation in CT without additional physician involvement, instead leveraging the cumulative work of radiologists over many years. 

Across both the hold-out test set and external test set, SAM2CT achieved the best performance both in terms of Dice score and lowest percent error in RECIST measurements. For the standard GSPS annotation prompts, line measurements achieve better segmentation results than arrow prompts. This is an expected result, as arrows can be significantly more ambiguous than line measurements and may even lead to the segmentation of incorrect objects in some cases.  

There are potentially several reasons why the SAM2 architecture is more suitable for this opportunistic segmentation task compared to UNet-based models. The first is the SAM2 approach benefits from large-scale pre-training , while the UNet-based models are trained from scratch. It is possible that extensive pre-training would lead to more accurate DynUNet and Swin UNETR models, but this requires a pre-training dataset. Additionally, UNet prompt-encoding approach (i.e., passing the prompts as a binary channel concatenated with the input) is likely hard for the model to learn due to the early fusing of the image channel and prompt channel within the model. On the other hand, SAM2 fully encodes the image and prompts separately and does not interpret them together until the mask decoder.

SAM2CT modifies the architecture of SAM2 such that instead of passing unconditioned image embeddings to the memory encoder, the memory encoder receives image embeddings after they have undergone memory attention. Across the hold-out test set and external test set, this MCM modification showed consistent benefits. In Fig.~\ref{fig:slice_dices}, we plot the 2D dice scores of individual slices as they get farther away from the original annotation slice. SAM2CT has similar performance compared to SAM (FT) on axial slices near the prompt slice. Instead, our MCM adaptation increases the quality of segmentation as the model gets further from the prompt slice. This is likely because MCM allows the model to capture more relevant features within the memory bank, rather than only tracking features related to a fixed number of slices. Interestingly, we showed that MCM was not only a successful fine-tuning strategy, but also a beneficial test-time adaptation for any SAM2-based model, without fine-tuning required. SAM2 (FT), MedSAM2, and SAM2 all increased segmentation performance when MCM was applied with no additional training. This suggests that MCM is inherently a more fitting architectural choice for 3D medical image segmentation. 

When tested on real-world physician-created GSPS objects from clinical PACS, SAM2CT was found to be highly useful for oncology, with only 13\% needing major adjustments to be clinically acceptable for lesion measurements. Moreover, SAM2CT had good generalization to cases with acute findings from the ED. While its training data did not include cases from the ED, SAM2CT is built upon a foundation model with good generalization capabilities. Many ED findings were accurately segmented, such as gallstones (Dice=0.725) and cysts (0.727). Unsurprisingly, SAM2CT does significantly better on ED with a pseudo-box label compared to the original physician-provided annotation (mostly lines and arrows), as box prompts inherently provide more information about an object. One challenge with ED cases is that physicians might use prompts differently for annotating ED cases than for oncology cases. For example, physicians often measure the neck of a hernia using a line rather than the hernia’s maximum diameter. Furthermore, some ED conditions were more challenging for the model, such as diverticulitis (Dice=0.302) and hematoma (0.528), due to the inherent difficulty in delineating boundaries for these pathologies. For example, SAM2CT was able to consistently segment the primary process of diverticulitis, but the model struggled to capture the entire inflamed segment.

Opportunistic promptable segmentation with SAM2CT has the potential to curate  large-scale multimodal datasets with paired 3D segmentation masks. Multimodal, grounded datasets may be obtained by linking SAM2CT-generated 3D masks with corresponding free-text radiology reports. This can be accomplished by matching findings from CT reports with characteristics of corresponding GSPS objects, including lesion diameter measurements. By aligning generated volumetric masks with semantically relevant report phrases, it becomes possible to construct large-scale vision–language datasets, facilitating the development of grounded report generation models in 3D CT.

There were several limitations to our study. The in-house ED and oncology test sets were relatively small and may not be representative of all institutions. Further external validation with different institutions’ data would be valuable to explore. Additionally, SAM2CT was trained exclusively on public CT segmentation datasets. These datasets are predominantly focused on organ and lesion segmentation and therefore may not capture the full spectrum of radiologic findings in routine clinical practice. Similarly, annotations were synthetically generated from ground truth segmentation masks during training, which may not accurately reflect the diversity of how radiologists place GSPS annotations in real-world settings . Lastly, despite SAM2CT’s promising performance on a diverse set of ED findings, our model was not trained on ED cases due to a lack of datasets with 3D contours. Future work will include expanding the training dataset of SAM2CT to cover a more diverse set of findings encountered in clinical CT imaging.

\section{\Large Conclusion}
\large
We developed and validated opportunistic promptable segmentation as a viable strategy for generating large-scale 3D CT segmentation datasets. Our promptable segmentation model, SAM2CT, was able to convert routine radiological annotations into 3D segmentation masks, achieving average Dice scores of 0.649 for arrow-based prompts and 0.757 for line-based prompts. It also performed well on real-world oncology cases and had promising generalizability to emergency department cases. Our model will allow researchers to leverage the extensive, underutilized annotation data already stored in clinical PACS systems to construct 3D CT segmentation datasets without additional manual radiologist effort.

\bibliography{sam2ct}

\begin{thebibliography}{34}
\providecommand{\natexlab}[1]{#1}
\providecommand{\url}[1]{\texttt{#1}}
\expandafter\ifx\csname urlstyle\endcsname\relax
  \providecommand{\doi}[1]{doi: #1}\else
  \providecommand{\doi}{doi: \begingroup \urlstyle{rm}\Url}\fi

\bibitem[Sivakumar et~al.(2025)Sivakumar, Lue, and Kundu]{fda_approval_jama}
R.~Sivakumar, B.~Lue, and S.~Kundu.
\newblock Fda approval of artificial intelligence and machine learning devices in radiology: A systematic review.
\newblock \emph{JAMA Netw Open}, 2025.
\newblock \doi{10.1001/jamanetworkopen.2025.42338}.

\bibitem[Singh et~al.(2025)Singh, Bapna, Diab, Ruiz, and Lotter]{fda_approval_npj}
R.~Singh, M.~Bapna, A.~R. Diab, E.~S. Ruiz, and W.~Lotter.
\newblock How ai is used in fda-authorized medical devices: a taxonomy across 1,016 authorizations.
\newblock \emph{npj Digit. Med.}, 2025.
\newblock \doi{10.1038/s41746-025-01800-1}.

\bibitem[Huemann et~al.(2025)Huemann, Church, Warner, Tran, Tie, McMillan, Hu, Cho, Lubner, and Bradshaw]{zach_visual_grounding}
Z.~Huemann, S.~Church, J.~D. Warner, D.~Tran, X.~Tie, A.~B. McMillan, J.~Hu, S.~Y. Cho, M.~Lubner, and T.~J. Bradshaw.
\newblock Vision-language modeling in pet/ct for visual grounding of positive findings.
\newblock \emph{arXiv}, 2025.
\newblock \doi{10.48550/arXiv.2502.00528}.

\bibitem[Sharma et~al.(2025)Sharma, Salvatelli, Srivastav, Bouzid, Bannur, Castro, Ilse, Bond-Taylor, Ranjit, Falck, Pérez-García, Schwaighofer, Richardson, Wetscherek, Hyland, and Alvarez-Valle]{maira_seg}
H.~Sharma, V.~Salvatelli, S.~Srivastav, K.~Bouzid, S.~Bannur, D.~C. Castro, M.~Ilse, S.~Bond-Taylor, M.~P. Ranjit, F.~Falck, F.~Pérez-García, A.~Schwaighofer, H.~Richardson, M.~Wetscherek, S.~Hyland, and J.~Alvarez-Valle.
\newblock Maira-seg: Enhancing radiology report generation with segmentation-aware multimodal large language models.
\newblock \emph{Proceedings of Machine Learning Research}, 259:\penalty0 941--960, 2025.

\bibitem[Gu et~al.(2024)Gu, Liu, Li, and Cai]{organ_mask_guided_gen}
T.~Gu, D.~Liu, Z.~Li, and W.~Cai.
\newblock Complex organ mask guided radiology report generation.
\newblock \emph{Proceedings of the IEEE/CVF Winter Conference on Applications of Computer Vision}, pages 7995--8004, 2024.

\bibitem[Zhou et~al.(2025)Zhou, Li, Li, Liu, Huang, Mao, Yang, Lv, and Liu]{tumor_volume_quantification}
Y.~Zhou, J.~Li, Q.~Li, L.~Liu, P.~Huang, Y.~Mao, Y.~Yang, F.~Lv, and Z.~Liu.
\newblock Ai-based quantification of enhancing tumor volume on contrast-enhanced mri to predict pathologic response and prognosis in hcc after haic plus targeted therapy and immunotherapy.
\newblock \emph{J Hepatocell Carcinoma}, 12:\penalty0 1509--1525, 2025.

\bibitem[Yin et~al.(2024)Yin, Li, Teng, Laghari, Almadhor, Gregus, and Sampedro]{brain_ct_classification}
S.~Yin, H.~Li, L.~Teng, A.~A. Laghari, A.~Almadhor, M.~Gregus, and G.~A. Sampedro.
\newblock Brain ct image classification based on mask rcnn and attention mechanism.
\newblock \emph{Sci Rep}, 2024.
\newblock \doi{10.1038/s41598-024-78566-1}.

\bibitem[Yoo et~al.(2025)Yoo, Namdar, Wagner, Yeom, Nobre, Tabori, Hawkins, Ertl-Wagner, and Khalvati]{gen_ai_4_tumor_classification}
J.~J. Yoo, K.~Namdar, M.~W. Wagner, K.~W. Yeom, L.~F. Nobre, U.~Tabori, C.~Hawkins, B.~B. Ertl-Wagner, and F.~Khalvati.
\newblock Generative ai for weakly supervised segmentation and downstream classification of brain tumors on mr images.
\newblock \emph{Sci Rep}, 2025.
\newblock \doi{10.1038/s41598-025-06741-z}.

\bibitem[Cao et~al.(2023)Cao, Xia, Yao, Han, Lambert, Zhang, Tang, Jin, Jiang, Fang, Nogues, Li, Guo, Wang, Fang, Qiu, Hou, Kovarnik, Vocka, Lu, Chen, Chen, Liu, Zhou, Xie, Zhang, Lu, Hager, Yuille, Lu, Shao, Shi, Zhang, Liang, Zhang, and Lu]{panc_cancer_detection}
K.~Cao, Y.~Xia, J.~Yao, X.~Han, L.~Lambert, T.~Zhang, W.~Tang, G.~Jin, H.~Jiang, X.~Fang, I.~Nogues, X.~Li, W.~Guo, Y.~Wang, W.~Fang, M.~Qiu, Y.~Hou, T.~Kovarnik, M.~Vocka, Y.~Lu, Y.~Chen, X.~Chen, Z.~Liu, J.~Zhou, C.~Xie, R.~Zhang, H.~Lu, G.~D. Hager, A.~L. Yuille, L.~Lu, C.~Shao, Y.~Shi, Q.~Zhang, T.~Liang, L.~Zhang, and J.~Lu.
\newblock Large-scale pancreatic cancer detection via non-contrast ct and deep learning.
\newblock \emph{Nat Med}, 29:\penalty0 3033--3043, 2023.

\bibitem[Xu et~al.(2025)Xu, Hosseini, Anderson, Rinaldi, Krishnan, Martel, and Goubran]{ssl_dino_3d}
T.~Xu, S.~Hosseini, C.~Anderson, A.~Rinaldi, R.~G. Krishnan, A.~L. Martel, and M.~Goubran.
\newblock A generalizable 3d framework and model for self-supervised learning in medical imaging.
\newblock \emph{npj Digit. Med.}, 2025.
\newblock \doi{10.1038/s41746-025-02035-w}.

\bibitem[Taher et~al.(2023)Taher, Ikuta, and Soni]{3d_ct_ssl}
M.~R.~H. Taher, M.~Ikuta, and R.~Soni.
\newblock Curriculum self-supervised learning for 3d ct cardiac image segmentation.
\newblock \emph{Proceedings of Machine Learning Research}, 225:\penalty0 145--156, 2023.

\bibitem[Li et~al.(2025)Li, Yuille, and Zhou]{3D_transfer}
W.~Li, A.~Yuille, and Z.~Zhou.
\newblock How well do supervised 3d models transfer to medical imaging tasks?
\newblock \emph{International Conference on Learning Representations}, 2025.
\newblock \doi{10.48550/arXiv.2501.11253}.

\bibitem[Ji et~al.(2022)Ji, Bai, Yang, Ge, Zhu, Zhang, Li, Zhang, Ma, Wan, and Luo]{amos_2022}
Y.~Ji, H.~Bai, J.~Yang, C.~Ge, Y.~Zhu, R.~Zhang, Z.~Li, L.~Zhang, W.~Ma, X.~Wan, and P.~Luo.
\newblock Amos: A large-scale abdominal multi-organ benchmark for versatile medical image segmentation.
\newblock \emph{Conference on Neural Information Processing Systems}, 2022.
\newblock \doi{10.48550/arXiv.2206.08023}.

\bibitem[Rister et~al.(2020)Rister, Yi, Shivakumar, Nobashi, and Rubin]{ct-org}
B.~Rister, D.~Yi, K.~Shivakumar, T.~Nobashi, and D.~L. Rubin.
\newblock Ct-org, a new dataset for multiple organ segmentation in computed tomography.
\newblock \emph{Sci Data}, 2020.
\newblock \doi{10.1038/s41597-020-00715-8}.

\bibitem[Luo et~al.(2022)Luo, Liao, Xiao, Chen, Song, Zhang, Li, Metaxas, Wang, and Zhang]{WORD-dataset}
X.~Luo, W.~Liao, J.~Xiao, J.~Chen, T.~Song, X.~Zhang, K.~Li, D.~N. Metaxas, G.~Wang, and S.~Zhang.
\newblock Word: A large scale dataset, benchmark and clinical applicable study for abdominal organ segmentation from ct image.
\newblock \emph{Medical Image Analysis}, 2022.
\newblock \doi{10.1016/j.media.2022.102642}.

\bibitem[Bilic et~al.(2022)Bilic, Christ, Li, Vorontsov, Ben-Cohen, Kaissis, Szeskin, Jacobs, Mamani, Chartrand, Lohöfer, Holch, Sommer, Hofmann, Hostettler, Lev-Cohain, Drozdzal, Amitai, Vivantik, Sosna, Ezhov, Sekuboyina, Navarro, Kofler, Paetzold, Shit, Hu, Lipková, Rempfler, Piraud, Kirschke, Wiestler, Zhang, Hülsemeyer, Beetz, Ettlinger, Antonelli, Bae, Bellver, Bi, Chen, Chlebus, Dam, Dou, Fu, Georgescu, i~Nieto, Gruen, Han, Heng, Hesser, Moltz, Igel, Isensee, Jäger, Jia, Kaluva, Khened, Kim, Kim, Kim, Kohl, Konopczynski, Kori, Krishnamurthi, Li, Li, Li, Li, Lowengrub, Ma, Maier-Hein, Maninis, Meine, Merhof, Pai, Perslev, Petersen, Pont-Tuset, Qi, Qi, Rippel, Roth, Sarasua, Schenk, Shen, Torres, Wachinger, Wang, Weninger, Wu, Xu, Yang, Yu, Yuan, Yu, Zhang, Cardoso, Bakas, Braren, Heinemann, Pal, Tang, Kadoury, Soler, van Ginneken, Greenspan, Joskowicz, and Menze]{lits_dataset}
P.~Bilic, P.~Christ, H.~B. Li, E.~Vorontsov, A.~Ben-Cohen, G.~Kaissis, A.~Szeskin, C.~Jacobs, G.~E.~H. Mamani, G.~Chartrand, F.~Lohöfer, J.~W. Holch, W.~Sommer, F.~Hofmann, A.~Hostettler, N.~Lev-Cohain, M.~Drozdzal, M.~M. Amitai, R.~Vivantik, J.~Sosna, I.~Ezhov, A.~Sekuboyina, F.~Navarro, F.~Kofler, J.~C. Paetzold, S.~Shit, X.~Hu, J.~Lipková, M.~Rempfler, M.~Piraud, J.~Kirschke, B.~Wiestler, Z.~Zhang, C.~Hülsemeyer, M.~Beetz, F.~Ettlinger, M.~Antonelli, W.~Bae, M.~Bellver, L.~Bi, H.~Chen, G.~Chlebus, E.~B. Dam, Q.~Dou, C.~Fu, B.~Georgescu, X.~Giró i~Nieto, F.~Gruen, X.~Han, P.~Heng, J.~Hesser, J.~H. Moltz, C.~Igel, F.~Isensee, P.~Jäger, F.~Jia, K.~C. Kaluva, M.~Khened, I.~Kim, J.~Kim, S.~Kim, S.~Kohl, T.~Konopczynski, A.~Kori, G.~Krishnamurthi, F.~Li, H.~Li, J.~Li, X.~Li, J.~Lowengrub, J.~Ma, K.~Maier-Hein, K.~Maninis, H.~Meine, D.~Merhof, A.~Pai, M.~Perslev, J.~Petersen, J.~Pont-Tuset, J.~Qi, X.~Qi, O.~Rippel, K.~Roth, I.~Sarasua, A.~Schenk, Z.~Shen, J.~Torres, C.~Wachinger, C.~Wang, L.~Weninger, J.~Wu,
  D.~Xu, X.~Yang, S.~Chun-Ho Yu, Y.~Yuan, M.~Yu, L.~Zhang, J.~Cardoso, S.~Bakas, R.~Braren, V.~Heinemann, C.~Pal, A.~Tang, S.~Kadoury, L.~Soler, B.~van Ginneken, H.~Greenspan, L.~Joskowicz, and B.~Menze.
\newblock The liver tumor segmentation benchmark (lits).
\newblock \emph{Medical Image Analysis}, 2022.
\newblock \doi{10.1016/j.media.2022.102680}.

\bibitem[Antonelli et~al.(2022)Antonelli, Reinke, Bakas, Farahani, Kopp-Schneider, Landman, Litjens, Menze, Ronneberger, M.Summers, van Ginneken, Bilello, Bilic, Christ, Do, Gollub, Heckers, Huisman, Jarnagin, McHugo, Napel, Pernicka, Rhode, Tobon-Gomez, Vorontsov, Huisman, Meakin, Ourselin, Wiesenfarth, Arbelaez, Bae, Chen, Daza, Feng, He, Isensee, Ji, Jia, Kim, Kim, Merhof, Pai, Park, Perslev, Rezaiifar, Rippel, Sarasua, Shen, Son, Wachinger, Wang, Wang, Xia, Xu, Xu, Zheng, Simpson, Maier-Hein, and Cardoso]{MSD-dataset}
M.~Antonelli, A.~Reinke, S.~Bakas, K.~Farahani, A.~Kopp-Schneider, B.~A. Landman, G.~Litjens, B.~Menze, O.~Ronneberger, R.~M.Summers, B.~van Ginneken, M.~Bilello, P.~Bilic, P.~F. Christ, R.~K.~G. Do, M.~J. Gollub, S.~H. Heckers, H.~Huisman, W.~R. Jarnagin, M.~K. McHugo, S.~Napel, J.~S.~G. Pernicka, K.~Rhode, C.~Tobon-Gomez, E.~Vorontsov, H.~Huisman, J.~A. Meakin, S.~Ourselin, M.~Wiesenfarth, P.~Arbelaez, B.~Bae, S.~Chen, L.~Daza, J.~Feng, B.~He, F.~Isensee, Y.~Ji, F.~Jia, N.~Kim, I.~Kim, D.~Merhof, A.~Pai, B.~Park, M.~Perslev, R.~Rezaiifar, O.~Rippel, I.~Sarasua, W.~Shen, J.~Son, C.~Wachinger, L.~Wang, Y.~Wang, Y.~Xia, D.~Xu, Z.~Xu, Y.~Zheng, A.~L. Simpson, L.~Maier-Hein, and M.~J. Cardoso.
\newblock The medical segmentation decathlon.
\newblock \emph{Nat Commun}, 2022.
\newblock \doi{10.1038/s41467-022-30695-9}.

\bibitem[Heller et~al.(2023)Heller, Isensee, Trofimova, Tejpaul, Zhao, Chen, Wang, Golts, Khapun, Shats, Shoshan, Gilboa-Solomon, George, Yang, Zhang, Zhang, Xia, Wu, Liu, Walczak, McSweeney, Vasdev, Hornung, Solaiman, Schoephoerster, Abernathy, Wu, Abdulkadir, Byun, Spriggs, Struyk, Austin, Simpson, Hagstrom, Virnig, French, Venkatesh, Chan, Moore, Jacobsen, Austin, Austin, Regmi, Papanikolopoulos, and Weight]{kits}
N.~Heller, F.~Isensee, D.~Trofimova, R.~Tejpaul, Z.~Zhao, H.~Chen, L.~Wang, A.~Golts, D.~Khapun, D.~Shats, Y.~Shoshan, F.~Gilboa-Solomon, Y.~George, X.~Yang, J.~Zhang, J.~Zhang, Y.~Xia, M.~Wu, Z.~Liu, E.~Walczak, S.~McSweeney, R.~Vasdev, C.~Hornung, R.~Solaiman, J.~Schoephoerster, B.~Abernathy, D.~Wu, S.~Abdulkadir, B.~Byun, J.~Spriggs, G.~Struyk, A.~Austin, B.~Simpson, M.~Hagstrom, S.~Virnig, J.~French, N.~Venkatesh, S.~Chan, K.~Moore, A.~Jacobsen, S.~Austin, M.~Austin, S.~Regmi, N.~Papanikolopoulos, and C.~Weight.
\newblock The kits21 challenge: Automatic segmentation of kidneys, renal tumors, and renal cysts in corticomedullary-phase ct.
\newblock \emph{arXiv}, 2023.
\newblock \doi{10.48550/arXiv.2307.01984}.

\bibitem[de~Grauw et~al.(2025)de~Grauw, Scholten, Smit, Rutten, Prokop, van Ginneken, and Hering]{uls23_dataset}
M.J.J. de~Grauw, E.~T. Scholten, E.~J. Smit, M.J.C.M. Rutten, M.~Prokop, B.~van Ginneken, and A.~Hering.
\newblock The uls23 challenge: A baseline model and benchmark dataset for 3d universal lesion segmentation in computed tomography.
\newblock \emph{Medical Image Analysis}, 2025.
\newblock \doi{10.1016/j.media.2025.103525}.

\bibitem[Roth et~al.(2014)Roth, Lu, Seff, Cherry, Hoffman, Wang, Liu, Turkbey, and Summers]{lymph_node_dataset}
H.R. Roth, L.~Lu, A.~Seff, K.M. Cherry, J.~Hoffman, S.~Wang, J.~Liu, E.~Turkbey, and R.M. Summers.
\newblock A new 2.5d representation for lymph node detection using random sets of deep convolutional neural network observations.
\newblock \emph{Med Image Comput Comput Assist Interv}, 17(Pt 1):\penalty0 520--527, 2014.

\bibitem[Kirillov et~al.(2023)Kirillov, Mintun, Ravi, Mao, Rolland, Gustafson, Xiao, Whitehead, Berg, Lo, Dollár, and Girshick]{sam_paper}
A.~Kirillov, E.~Mintun, N.~Ravi, H.~Mao, C.~Rolland, L.~Gustafson, T.~Xiao, S.~Whitehead, A.~C. Berg, W.~Lo, P.~Dollár, and R.~Girshick.
\newblock Segment anything.
\newblock \emph{arXiv}, 2023.
\newblock \doi{10.48550/arXiv.2304.02643}.

\bibitem[Ma et~al.(2024)Ma, He, Li, Han, You, and Wang]{medsam}
J.~Ma, Y.~He, F.~Li, L.~Han, C.~You, and B.~Wang.
\newblock Segment anything in medical images.
\newblock \emph{Nat Commun}, 2024.
\newblock \doi{10.1038/s41467-024-44824-z}.

\bibitem[Wang et~al.(2024)Wang, Guo, Ye, Deng, Cheng, Li, Chen, Su, Huang, Shen, Fu, Zhang, He, and Qiao]{sam_med_3d}
H.~Wang, S.~Guo, J.~Ye, Z.~Deng, J.~Cheng, T.~Li, J.~Chen, Y.~Su, Z.~Huang, Y.~Shen, B.~Fu, S.~Zhang, J.~He, and Y.~Qiao.
\newblock Sam-med3d: Towards general-purpose segmentation models for volumetric medical images.
\newblock \emph{arXiv}, 2024.
\newblock \doi{10.48550/arXiv.2310.15161}.

\bibitem[Bui et~al.(2024)Bui, Hoang, Tran, Doretto, Adjeroh, Patel, Choudhary, and Le]{sam3d}
N.~Bui, D.~Hoang, M.~Tran, G.~Doretto, D.~Adjeroh, B.~Patel, A.~Choudhary, and N.~Le.
\newblock Sam3d: Segment anything model in volumetric medical images.
\newblock \emph{International Symposium on Biomedical Imaging}, 2024.
\newblock \doi{10.48550/arXiv.2309.03493}.

\bibitem[Ravi et~al.(2024)Ravi, Gabeur, Hu, Hu, Ryali, Ma, Khedr, Rädle, Rolland, Gustafson, Mintun, Pan, Alwala, Carion, Wu, Girshick, Dollár, and Feichtenhofer]{sam2}
N.~Ravi, V.~Gabeur, Y.~Hu, R.~Hu, C.~Ryali, T.~Ma, H.~Khedr, R.~Rädle, C.~Rolland, L.~Gustafson, E.~Mintun, J.~Pan, K.~V. Alwala, N.~Carion, C.~Wu, R.~Girshick, P.~Dollár, and C.~Feichtenhofer.
\newblock Sam 2: Segment anything in images and videos.
\newblock \emph{arXiv}, 2024.
\newblock \doi{10.48550/arXiv.2408.00714}.

\bibitem[Zhu et~al.(2024)Zhu, Hamdi, Qi, Jin, and Wu]{medical_sam_2}
Jiayuan Zhu, Abdullah Hamdi, Yunli Qi, Yueming Jin, and Junde Wu.
\newblock Medical sam 2: Segment medical images as video via segment anything model 2.
\newblock \emph{arXiv}, 2024.
\newblock \doi{10.48550/arXiv.2408.00874}.

\bibitem[Ma et~al.(2025)Ma, Yang, Kim, Chen, Baharoon, Fallahpour, Asakereh, Lyu, and Wang]{MedSAM2}
Jun Ma, Zongxin Yang, Sumin Kim, Bihui Chen, Mohammed Baharoon, Adibvafa Fallahpour, Reza Asakereh, Hongwei Lyu, and Bo~Wang.
\newblock Medsam2: Segment anything in 3d medical images and videos.
\newblock \emph{arXiv}, 2025.
\newblock \doi{10.48550/arXiv.2504.03600}.

\bibitem[Yan et~al.(2018)Yan, Wang, Lu, and Summers]{deeplesion}
Ke~Yan, Xiaosong Wang, Le~Lu, and Ronald~M. Summers.
\newblock Deeplesion: automated mining of large-scale lesion annotations and universal lesion detection with deep learning.
\newblock \emph{J Med Imaging (Bellingham)}, 2018.
\newblock \doi{10.1117/1.JMI.5.3.036501}.

\bibitem[Jain et~al.(2021)Jain, Agrawal, Saporta, Truong, Duong, Bui, Chambon, Zhang, Lungren, Ng, Langlotz, and Rajpurkar]{radgraph}
Saahil Jain, Ashwin Agrawal, Adriel Saporta, Steven~QH Truong, Du~Nguyen Duong, Tan Bui, Pierre Chambon, Yuhao Zhang, Matthew~P. Lungren, Andrew~Y. Ng, Curtis~P. Langlotz, and Pranav Rajpurkar.
\newblock Radgraph: Extracting clinical entities and relations from radiology reports.
\newblock \emph{Conference on Neural Information Processing Systems}, 2021.
\newblock \doi{10.48550/arXiv.2106.14463}.

\bibitem[Eisenhauer et~al.(2023)Eisenhauer, Therasse, Bogaerts, Schwartz, Sargent, Ford, Dancey, Arbuck, Gwyther, Mooney, Rubinstein, Shankar, Dodd, Kaplan, Lacombe, and Verweij]{hiera}
E.~A. Eisenhauer, P.~Therasse, J.~Bogaerts, L.~H. Schwartz, D.~Sargent, R.~Ford, J.~Dancey, S.~Arbuck, S.~Gwyther, M.~Mooney, L.~Rubinstein, L.~Shankar, L.~Dodd, R.~Kaplan, D.~Lacombe, and J.~Verweij.
\newblock Hiera: A hierarchical vision transformer without the bells-and-whistles.
\newblock \emph{International Conference on Machine Learning}, 2023.
\newblock \doi{10.48550/arXiv.2306.00989}.

\bibitem[Isensee et~al.(2021)Isensee, Petersen, Klein, Zimmerer, Jaeger, Kohl, Wasserthal, Koehler, Norajitra, Wirkert, and Maier-Hein]{nnunet}
Fabian Isensee, Jens Petersen, Andre Klein, David Zimmerer, Paul~F. Jaeger, Simon Kohl, Jakob Wasserthal, Gregor Koehler, Tobias Norajitra, Sebastian Wirkert, and Klaus~H. Maier-Hein.
\newblock nnu-net: a self-configuring method for deep learning-based biomedical image segmentation.
\newblock \emph{Nat Methods}, 18:\penalty0 203--211, 2021.

\bibitem[Cardoso et~al.(2022)Cardoso, Li, Brown, Ma, Kerfoot, Wang, Murrey, Myronenko, Zhao, Yang, Nath, He, Xu, Hatamizadeh, Myronenko, Zhu, Liu, Zheng, Tang, Yang, Zephyr, Hashemian, Alle, Darestani, Budd, Modat, Vercauteren, Wang, Li, Hu, Fu, Gorman, Johnson, Genereaux, Erdal, Gupta, Diaz-Pinto, Dourson, Maier-Hein, Jaeger, Baumgartner, Kalpathy-Cramer, Flores, Kirby, Cooper, Roth, Xu, Bericat, Floca, Zhou, Shuaib, Farahani, Maier-Hein, Aylward, Dogra, Ourselin, and Feng]{monai}
M.~Jorge Cardoso, Wenqi Li, Richard Brown, Nic Ma, Eric Kerfoot, Yiheng Wang, Benjamin Murrey, Andriy Myronenko, Can Zhao, Dong Yang, Vishwesh Nath, Yufan He, Ziyue Xu, Ali Hatamizadeh, Andriy Myronenko, Wentao Zhu, Yun Liu, Mingxin Zheng, Yucheng Tang, Isaac Yang, Michael Zephyr, Behrooz Hashemian, Sachidanand Alle, Mohammad~Zalbagi Darestani, Charlie Budd, Marc Modat, Tom Vercauteren, Guotai Wang, Yiwen Li, Yipeng Hu, Yunguan Fu, Benjamin Gorman, Hans Johnson, Brad Genereaux, Barbaros~S. Erdal, Vikash Gupta, Andres Diaz-Pinto, Andre Dourson, Lena Maier-Hein, Paul~F. Jaeger, Michael Baumgartner, Jayashree Kalpathy-Cramer, Mona Flores, Justin Kirby, Lee A.~D. Cooper, Holger~R. Roth, Daguang Xu, David Bericat, Ralf Floca, S.~Kevin Zhou, Haris Shuaib, Keyvan Farahani, Klaus~H. Maier-Hein, Stephen Aylward, Prerna Dogra, Sebastien Ourselin, and Andrew Feng.
\newblock Monai: An open-source framework for deep learning in healthcare.
\newblock \emph{arXiv}, 2022.
\newblock \doi{10.48550/arXiv.2211.02701}.

\bibitem[Hatamizadeh et~al.(2022)Hatamizadeh, Nath, Tang, Yang, Roth, and Xu]{swin_unetr}
Ali Hatamizadeh, Vishwesh Nath, Yucheng Tang, Dong Yang, Holger Roth, and Daguang Xu.
\newblock Swin unetr: Swin transformers for semantic segmentation of brain tumors in mri images.
\newblock \emph{Brainlesion: Glioma, Multiple Sclerosis, Stroke and Traumatic Brain Injuries}, 7:\penalty0 272--–284, 2022.

\bibitem[Eisenhauer et~al.(2009)Eisenhauer, Therasse, Bogaerts, Schwartz, Sargent, Ford, Dancey, Arbuck, Gwyther, Mooney, Rubinstein, Shankar, Dodd, Kaplan, Lacombe, and Verweij]{recist}
E.~A. Eisenhauer, P.~Therasse, J.~Bogaerts, L.~H. Schwartz, D.~Sargent, R.~Ford, J.~Dancey, S.~Arbuck, S.~Gwyther, M.~Mooney, L.~Rubinstein, L.~Shankar, L.~Dodd, R.~Kaplan, D.~Lacombe, and J.~Verweij.
\newblock New response evaluation criteria in solid tumours: revised recist guideline (version 1.1).
\newblock \emph{Eur J Cancer}, 45:\penalty0 228--247, 2009.

\end{thebibliography}

\label{lastpage}
\end{document}